\documentclass[10pt,twocolumn,letterpaper]{article}

\usepackage{cvpr}              

\usepackage{multirow}
\usepackage{graphicx}
\usepackage{times}
\usepackage{epsfig}
\usepackage{amsmath}
\usepackage{amssymb}
\usepackage{xcolor}
\usepackage{comment}
\usepackage{bbm}

\definecolor{cvprblue}{rgb}{0.21,0.49,0.74}
\usepackage[pagebackref,breaklinks,colorlinks,allcolors=cvprblue]{hyperref}

\def\paperID{519} 
\def\confName{CVPR}
\def\confYear{2025}

\title{PanoSLAM: Panoptic 3D Scene Reconstruction via Gaussian SLAM}
\author{Runnan Chen$^{1}$~\quad Zhaoqing Wang$^{1}$~\quad Jiepeng Wang$^{2}$ \\  Yuexin Ma$^{3}$ ~\quad Mingming Gong$^{4}$ ~\quad Wenping Wang$^{5}$ ~\quad Tongliang Liu$^{1}$
\\[0.2ex]
\small{$^{1}$The University of Sydney} \quad 
\small{$^{2}$The University of Hong Kong} \\ 
\small{$^{3}$ShanghaiTech University} \quad
\small{$^{4}$The University of Melbourne} \quad
\small{$^{5}$Texas A\&M University}
}

\begin{document}
\maketitle

\begin{abstract}
Understanding geometric, semantic, and instance information in 3D scenes from sequential video data is essential for applications in robotics and augmented reality. However, existing Simultaneous Localization and Mapping (SLAM) methods generally focus on either geometric or semantic reconstruction. In this paper, we introduce PanoSLAM, the first SLAM system to integrate geometric reconstruction, 3D semantic segmentation, and 3D instance segmentation within a unified framework. Our approach builds upon 3D Gaussian Splatting, modified with several critical components to enable efficient rendering of depth, color, semantic, and instance information from arbitrary viewpoints. To achieve panoptic 3D scene reconstruction from sequential RGB-D videos, we propose an online Spatial-Temporal Lifting (STL) module that transfers 2D panoptic predictions from vision models into 3D Gaussian representations. This STL module addresses the challenges of label noise and inconsistencies in 2D predictions by refining the pseudo labels across multi-view inputs, creating a coherent 3D representation that enhances segmentation accuracy. Our experiments show that PanoSLAM outperforms recent semantic SLAM methods in both mapping and tracking accuracy. For the first time, it achieves panoptic 3D reconstruction of open-world environments directly from the RGB-D video. \footnote{\url{https://github.com/runnanchen/PanoSLAM}.}

\end{abstract}

\section{Introduction}
\label{sec:introduction}

\begin{figure}
  \centerline{\includegraphics[width=0.5\textwidth]{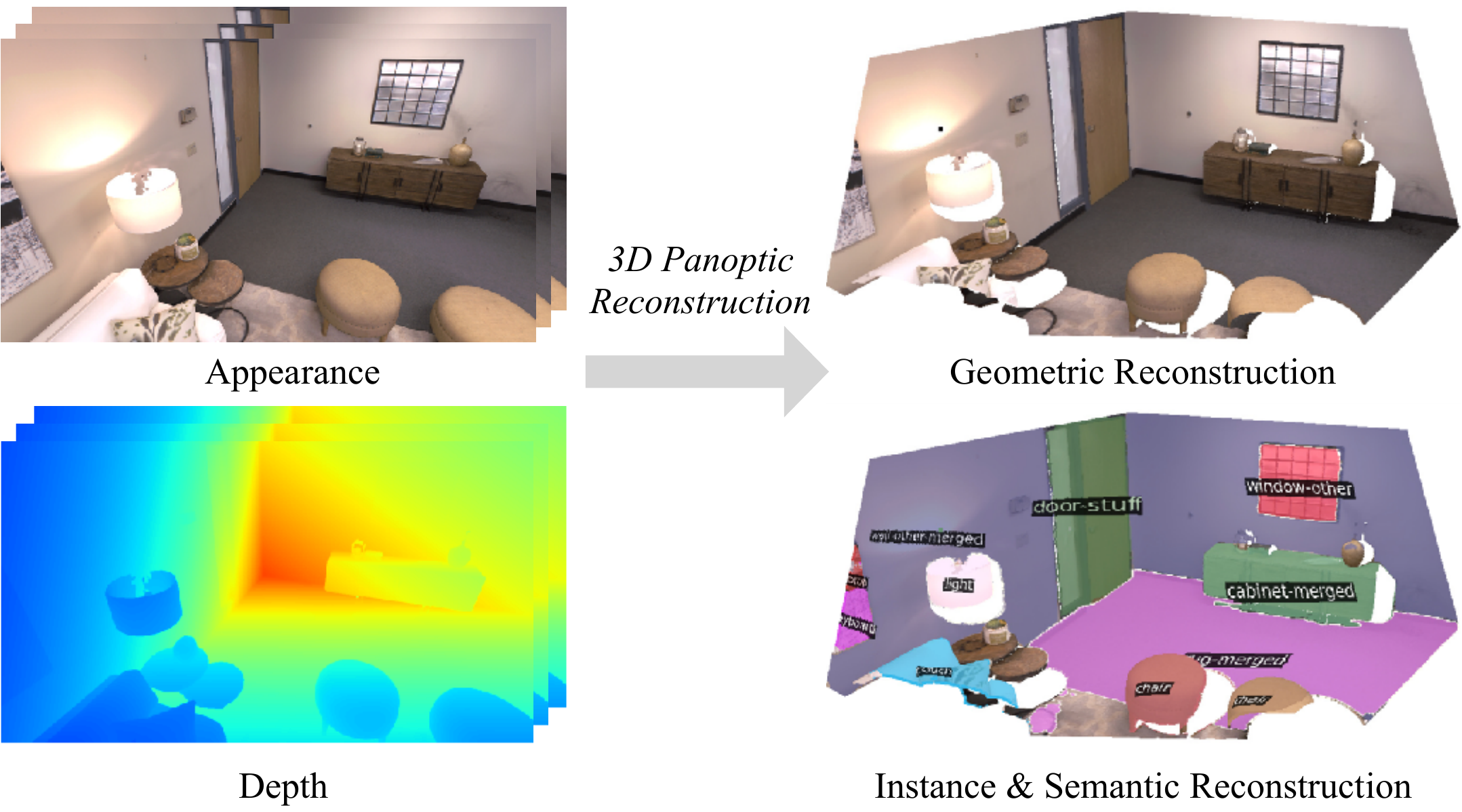}}
  \caption{We present PanoSLAM, a SLAM system based on 3D Gaussian Splatting, capable of 3D geometric, semantic and instance reconstruction from unlabeled RGB-D videos.}
  \label{fig:teaser}
\end{figure}

Semantic Simultaneous Localization and Mapping (SLAM) combines scene reconstruction and camera pose estimation with semantic scene understanding, providing a more comprehensive interpretation of the environment than traditional SLAM. By generating 3D semantic maps, semantic SLAM facilitates applications in diverse fields, including autonomous driving, robotic navigation, and digital city planning. These semantic maps contribute to advanced decision-making and environment interaction, making semantic SLAM a cornerstone technology in intelligent systems.

However, existing semantic SLAM methods \cite{zhu2023sni,li2023dns,haghighi2023neural,li2024sgs,zhu2024semgauss} face limitations in fully capturing the panoptic nature of 3D scenes, an essential aspect that includes both instance and semantic-level details. Additionally, these methods generally rely on densely labeled scenes for semantic mapping, a requirement that is labor-intensive, time-consuming, and cost-prohibitive, especially in open-world environments where scene diversity makes manual labeling impractical. This challenge underscores a fundamental question: How can we reconstruct a panoptic 3D scene from sequential video data without the need for manual semantic annotation?

Our approach is inspired by recent breakthroughs in vision foundation models such as CLIP \cite{radford2021learning} and SAM \cite{kirillov2023segment}, which demonstrate remarkable zero-shot perception capabilities across diverse environments. These foundation models have opened up new possibilities for transferring 2D vision knowledge to 3D representations, which include point clouds \cite{chen2024towards,chen2023clip2scene,peng2022openscene}, Neural Fields \cite{kerr2023lerf,siddiqui2023panoptic}, and 3D Gaussians \cite{shi2023language,qin2024langsplat}, enabling label-free 3D scene understanding. However, the application of these models in SLAM has been hindered by substantial offline optimization requirements, which conflict with the online demands of SLAM.

Recent advancements in Gaussian Splatting \cite{kerbl20233d} have shown promising results for scene reconstruction by leveraging 3D Gaussian representations, which allow for high-quality and efficient rendering through a splatting-based approach. Some SLAM systems \cite{keetha2023splatam,matsuki2023gaussian,yan2023gs,yugay2023gaussian} have adopted 3D Gaussian Splatting to achieve photorealistic scene mapping. Yet, integrating vision foundation models into SLAM systems for open-world 3D scene understanding remains an unexplored area, leaving a gap in fully panoptic SLAM reconstruction methods.

In this work, we address this gap by introducing \textbf{PanoSLAM}, a novel SLAM framework that enables panoptic 3D scene reconstruction from unlabeled RGB-D video input. Our method builds upon the Gaussian Splatting technique, enhanced with critical modifications for panoptic rendering, including semantic Gaussian initialization, densification, and panoptic segmentation formulation. A primary challenge in our approach is dealing with label noise, as we rely on pseudo-labels from 2D panoptic predictions provided by vision models. These pseudo-labels are susceptible to noise, such as inconsistencies in mask predictions and class labels across different views, which can create conflicts during optimization and degrade semantic map quality. To tackle this challenge, we introduce a Spatial-Temporal Lifting (STL) module that refines the noisy pseudo-labels by projecting them into 3D space, leveraging multi-view consistency to enhance the reliability of labels in 3D. Our STL module integrates multi-view 2D panoptic predictions to create a cohesive 3D representation, addressing label noise and facilitating high-quality panoptic scene reconstruction.

We evaluate PanoSLAM on benchmark datasets, including Replica \cite{straub2019replica} and ScanNet++ \cite{yeshwanth2023scannet++}, where our method significantly outperforms recent semantic SLAM approaches in mapping and tracking accuracy. Notably, PanoSLAM is the first method to achieve panoptic 3D scene reconstruction without manual labels. Our work combines efficient Gaussian Splatting with vision foundation models, to extend the possibilities of panoptic 3D reconstruction in diverse open-world environments.

In summary, our contributions are as follows:
\begin{itemize}
    \item We introduce the first panoptic 3D scene reconstruction method based on Gaussian Splatting within a SLAM framework.
    \item We propose an innovative Spatial-Temporal Lifting module for consistent 2D-to-3D knowledge distillation across multiple views, addressing challenges with noisy labels in panoptic reconstruction.
    \item Our experimental results demonstrate that PanoSLAM achieves state-of-the-art performance, pioneering label-free panoptic 3D scene reconstruction.
\end{itemize}

\section{Related Work}
\label{sec:relatedwork}
\paragraph{Scene Understanding.}
Scene understanding, which focuses on recognizing objects and their relationships, is essential in fields like robotics, autonomous driving, and smart cities. While supervised methods have significantly advanced both 2D and 3D scene understanding~\cite{zhu2020cylindrical, wu2022point, cheng2020panoptic, qi2017pointnet, zhu2021cylindrical, rpvnet, af2s3net, kong2023rethinking, hong2022dsnet, cheng2021per, strudel2021segmenter, cheng2022masked,sun2024empirical,yin2024fusion,xu2023human,kong2023robo3d,contributors2020mmdetection3d,liu2023uniseg}, they rely on extensive annotations, limiting their adaptability to new object categories outside the training data. To address these limitations, some approaches focus on open-world scene understanding~\cite{chen2023bridging,chen2022zero, michele2021generative, ding2022language, riz2023novel, xu2021simple, lilanguage, bucher2019zero, li2020consistent, hu2020uncertainty, zhang2021prototypical,liu2024multi,chen2023bridging,lu2023see,chen2023model2scene}, while others~\cite{peng2023openscene, XuYan20222DPASS2P, chen2023clip2scene, sautier2022image, chen2024towards, peng2025learning} aim to reduce 3D annotation demands by leveraging knowledge from 2D networks. However, these methods often face challenges in open-world scenarios where label-free, real-time understanding is crucial. Vision foundation models like CLIP~\cite{radford2021learning} and SAM~\cite{kirillov2023segment} have shown strong potential in open-world tasks. Efforts like CLIP2Scene~\cite{chen2023clip2scene} and CNS~\cite{chen2024towards} transfer 2D vision model knowledge to 3D representations (e.g., point clouds, Neural Fields, and 3D Gaussians) for label-free 3D understanding, often using a 2D-3D calibration matrix. However, these approaches typically require extensive offline optimization, which restricts their online application in SLAM. Our work integrates online SLAM with knowledge from 2D vision models, enabling efficient reconstruction of 3D panoptic semantic maps from unlabeled RGB-D videos.

\vspace*{-3ex}
\paragraph{Semantic SLAM.}
Traditional SLAM methods follow frameworks like MonoSLAM~\cite{davison2007monoslam}, PTAM~\cite{klein2007parallel}, and ORB SLAM~\cite{mur2015orb}, which separate tasks into mapping and tracking. Recent approaches leverage neural implicit representation and rendering for dense SLAM~\cite{zhu2023sni,sucar2021imap,zhu2022nice,johari2023eslam,sandstrom2023point,wang2023co}. Gaussian Splatting~\cite{kerbl20233d}, based on 3D Gaussians, has emerged as a more efficient alternative for 3D scene reconstruction~\cite{kerbl20233d,keselman2022approximate,keselman2023flexible,wang2022voge,li2024urban4d}, inspiring Gaussian-based SLAM methods~\cite{keetha2023splatam,matsuki2023gaussian,yan2023gs,yugay2023gaussian}. For example, SplaTAM~\cite{keetha2023splatam} uses silhouette-guided optimization for dense mapping, while Gaussian SLAM~\cite{matsuki2023gaussian} incorporates Gaussian insertion and pruning for monocular SLAM.

Semantic SLAM methods~\cite{zhu2023sni,li2023dns,haghighi2023neural,li2024sgs,zhu2024semgauss} enhance environmental understanding by integrating semantic information into SLAM. Object-aware systems like SLAM++~\cite{salas2013slam} and Kimera~\cite{rosinol2020kimera} capture object-level or semantic mesh information, while recent methods such as SGS-SLAM~\cite{li2024sgs} and SemGauss-SLAM~\cite{zhu2024semgauss} apply 3D Gaussian Splatting for semantic SLAM. However, these methods depend on manual labels and struggle with open-world generalization, limiting their scalability and adaptability to diverse, unstructured environments. In contrast, this paper proposes a Gaussian-based SLAM system that reconstructs 3D panoptic semantic maps in open-world environments without any input labels, achieving robust, label-free scene understanding suitable for online applications.

\begin{figure*}
  \centerline{\includegraphics[width=1\textwidth]{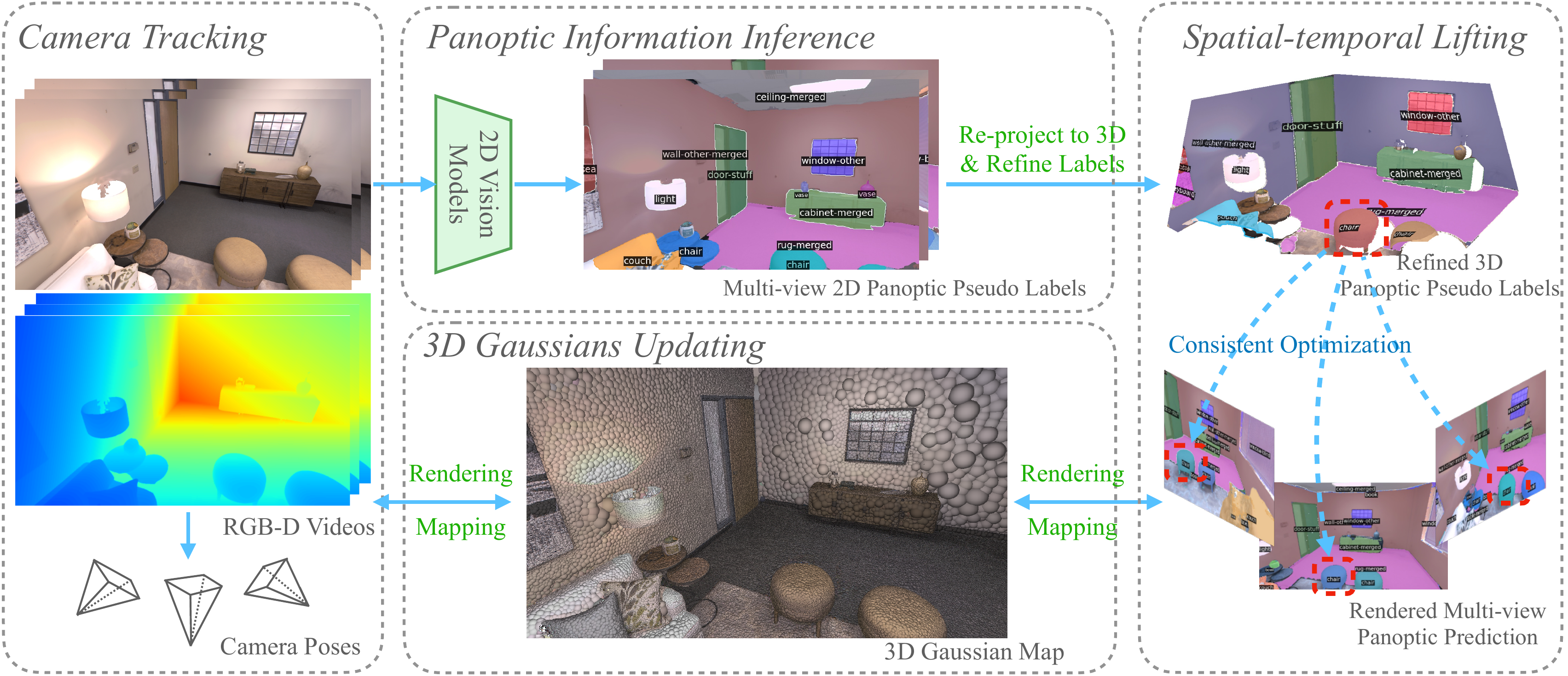}}
  \vspace{-1.5ex}
  \caption{Overview of the PanoSLAM framework for panoptic 3D scene reconstruction from unlabeled RGB-D videos. The system comprises four main components: (1) Camera Tracking: Estimating camera poses from input RGB-D videos to facilitate 3D mapping. (2) Panoptic Information Inference: Utilizing 2D vision models to generate multi-view 2D panoptic pseudo-labels, which are re-projected into 3D space for label refinement. (3) 3D Gaussians Updating: Constructing and updating a 3D Gaussian map for efficient rendering and mapping. (4) Spatial-Temporal Lifting: Refining 3D panoptic pseudo-labels through consistent optimization across multiple views to address label noise, enabling high-quality panoptic scene reconstruction.}
  \label{fig:framework}
\end{figure*}

\section{Methodology}
\label{sec:methodology}
Current semantic SLAM methods fall short in capturing both instance-level and semantic details, collectively known as panoptic information. In this section, we introduce a new method, called \textit{PanoSLAM}, designed to efficiently reconstruct 3D panoptic semantic maps from unlabeled RGB-D videos. PanoSLAM achieves this by transferring the knowledge of a 2D vision model into an online SLAM system. Specifically, we enhance a Gaussian-based SLAM framework with the capability to render panoptic information through targeted modifications to the Gaussian Splatting-based SLAM. Additionally, we develop a novel Spatial-Temporal Lifting module to handle noisy labels effectively. The following sections provide a brief overview of the Gaussian-based SLAM system, detail our modifications, and present the Spatial-Temporal Lifting module.

\subsection{Preliminaries about Gaussian-based SLAM}
\vspace*{-1ex}
Our system builds upon SplaTAM~\cite{keetha2023splatam}, a state-of-the-art dense RGB-D SLAM solution that leverages 3D Gaussian Splatting. SplaTAM represents the environment as a collection of 3D Gaussians, which can be rendered into high-quality color and depth images. By utilizing differentiable rendering and gradient-based optimization, it jointly optimizes the camera pose for each frame and constructs a volumetric map of the scene. In the following sections, we provide a brief overview of 3D Gaussian Splatting, followed by a detailed description of each module in SplaTAM.
\vspace*{-1ex}
\paragraph{Gaussian Representation.}
SplaTAM simplifies the Gaussian representation by adopting view-independent color and ensuring isotropy of the Gaussians. Each Gaussian is thus characterized by only eight parameters: three values for its RGB color \(\mathbf{c} \in \mathbb{R}^3\), three for its center position \(\mathbf{u} \in \mathbb{R}^3\), one for the radius (standard deviation) \(r\), and one for opacity \(o\). The contribution of each Gaussian to a point in 3D space \(\mathbf{x} \in \mathbb{R}^3\) is determined by the standard (unnormalized) Gaussian function, modulated by its opacity:
\begin{equation}\label{equ:Gaussian}
f(x)=o\exp \left(-\frac{\parallel \mathbf{x}-\mathbf{u}\parallel^2}{2r^2}\right).
\end{equation}
\vspace*{-1ex}
\paragraph{Differentiable Rendering via Splatting.}
SplaTAM renders an RGB image by sorting a collection of 3D Gaussians from front to back and then efficiently alpha-compositing the splatted 2D projection of each Gaussian in pixel space. The color of a rendered pixel $P = (u,v)$ can be expressed as:
\begin{equation}\label{equ:Rendering}
C(P)=\sum_{i=1}^{n}c_if_i(P)\prod_{j=1}^{i-1}(1-f_{i}(P)),
\end{equation}
where $f_i(P)$ is calculated according to Eq. \ref{equ:Gaussian}, using the $\mathbf{u}$ and $r$ values of the splatted 2D Gaussians in pixel space: 
\begin{equation}\label{equ:2DGaussian}
\mathbf{u}^{2D}=G\frac{E_t \mathbf{u}}{d},\;\;\; r^{2D}=\frac{fr}{d},\;\;\; d=(E_t \mathbf{u})_z,
\end{equation}
where the variables $G$, $E_t$, $f$, and $d$ represent the camera intrinsic matrix, extrinsic matrix for the rotation and translation of the camera at frame $t$, focal length (known), and the depth of the $i$-th Gaussian in camera coordinates, respectively.

\subsection{PanoSLAM}
Our PanoSLAM method, built upon Gaussian-based SLAM, integrates 2D vision model knowledge into a Gaussian SLAM system to efficiently reconstruct 3D panoptic maps from unlabeled RGB-D videos. To enable this, we implement key modifications for Splatting-based SLAM, allowing it to render panoptic information through semantic Gaussian initialization, densification, and panoptic segmentation formulation. The system leverages the 2D vision model’s panoptic predictions, including instance and semantic labels, as pseudo-labels to guide 3D panoptic map reconstruction. However, these pseudo-labels are susceptible to label noise, such as inconsistent mask and class predictions across views, which can create optimization conflicts and degrade the quality of the semantic map. To mitigate the effects of noisy pseudo-labels, we introduce a Spatial-Temporal Lifting module. In the following, we explain the steps and insights of PanoSLAM in detail.

\vspace*{-1ex}
\paragraph{Semantic Gaussian Representation, Initialization, and Densification.}
In contrast to SplaTAM, where each Gaussian is parameterized with 8 values, PanoSLAM assigns 13 values per Gaussian to facilitate semantic rendering. These include three values for its semantic embedding $\mathbf{s}\in \mathbb{R}^3$, one for its semantic radius $\hat{r}$ (standard deviation), and one for semantic opacity $\hat{o}$, along with the same 8 parameters as in SplaTAM. The semantic embedding of a pixel $P = (u,v)$ is thus rendered as:
\begin{equation}\label{equ:Rendering_semantics}
S(P)=\sum_{i=1}^{n}\mathbf{s}_i\hat{f}_i(P)\prod_{j=1}^{i-1}(1-\hat{f}_{i}(P)),
\end{equation}
where $\hat{f}_i(P)$ is calculated using Eqs. \ref{equ:Gaussian} and \ref{equ:2DGaussian} with $\hat{r}$ and $\hat{o}$ values.

To initialize the Gaussians in the first frame, we set the camera pose to identity and add a new Gaussian for each pixel. The center $\mathbf{u}$ is determined by unprojecting the pixel depth, and both the radius $r$ and semantic radius $\hat{r}$ are set to a one-pixel radius in the 2D image, i.e., $\hat{r}=r=\frac{D_{GT}}{f}$. The color $\mathbf{c}$ and semantic embedding $\mathbf{s}$ are initialized to the pixel color, while both optical opacity $o$ and semantic opacity $\hat{o}$ are initialized to 0.5.

For densification in subsequent frames, we add Gaussian kernels at locations where existing ones fail to adequately capture the scene's geometry or semantics. To identify these areas, we define a densification mask as follows:
\begin{equation}\label{equ:densification}
\begin{aligned}
    M(P)=(F(P) < 0.5)\; {\rm or} \; (\hat{F}(P) < 0.5) \; {\rm or} 
    \\
    (L(D(P))>T),
\end{aligned}
\end{equation}
where $D(P)$ is the differentiably rendered depth:
\begin{equation}\label{equ:depth}
D(P)=\sum_{i=1}^{n}d_if_i(P)\prod_{j=1}^{i-1}(1-f_{i}(P)),
\end{equation}
and $\hat{F}(P)$ is the silhouette map for visibility determination, calculated using $\hat{o}$ and $\hat{r}$:
\begin{equation}\label{equ:silhouette}
\hat{F}(P)=\sum_{i=1}^{n}\hat{f}_i(P)\prod_{j=1}^{i-1}(1-\hat{f}_{i}(P)),
\end{equation}
where $T$ is a threshold set to 50 times the median depth error. The densification mask serves a dual purpose: identifying areas where the map lacks density ($S < 0.5$) in geometry or semantics and detecting discrepancies between the ground-truth depth and the rendered depth. For each pixel flagged by this mask, we introduce a new Gaussian using the same initialization as in the first frame.

\vspace*{-1ex}
\paragraph{Camera Tracking.}
Camera tracking involves estimating the camera pose of the current incoming online RGB-D image in the input stream. The camera pose is initialized for a new timestep through a constant velocity forward projection of the pose parameters in the camera centre + quaternion space:
\begin{equation}\label{equ:camera_pose}
E_{t+1}=E_t+(E_t-E_{t-1}).
\end{equation}
The camera pose $E_t$ is then iteratively refined through gradient-based optimization by differentiably rendering RGB, depth, and silhouette maps and adjusting the camera parameters to minimize the loss, while fixing Gaussian parameters.

\vspace*{-1ex}
\paragraph{Panoptic Segmentation Formulation.}
Panoptic segmentation is achieved by combining semantic and instance segmentation. Inspired by Maskformer \cite{cheng2021per}, we decompose the panoptic segmentation task into two steps: 1) segmenting the image into $N$ regions using binary masks, and 2) assigning each region a probability distribution over $K$ categories. Specifically, for each pixel $P = (u,v)$, we first generate region predictions:
\begin{equation}\label{equ:region_prediction}
R(P)=\Gamma(S(P)) \otimes \mathbb{M},
\end{equation}
where $R(P) \in \mathbb{R}^N$ represents the predicted distribution over $N$ regions, $\mathbb{M} \in \mathbb{R}^{N \times H}$ is a set of $N$ region embeddings with $H$ dimensions, $\Gamma(\bullet)$ is an MLP decoder that elevates the semantic embedding dimension from 3 to $H$, and $\otimes$ denotes matrix multiplication. The $N$ regions are then classified into $K$ categories as follows:
\begin{equation}\label{equ:region_classfication}
O(\mathbb{M})=\mathbb{M} \otimes \mathbb{C},
\end{equation}
where $\mathbb{C} \in \mathbb{R}^{N \times K}$ serves as the classifier, and $O(\mathbb{M}) \in \mathbb{R}^K$ represents the predicted distribution over $K$ categories. These definitions effectively assign a pixel at location $P = (u,v)$ to the $i$-th probability-region and $j$-th probability-category only if both the region prediction probability $R^i(P)$ and the highest class probability $O^j(\mathbb{M})$ are sufficiently high~\cite{cheng2021per}. Low-confidence predictions are discarded before inference, and segments with substantial parts of their binary masks occluded by other predictions are also removed. 
\vspace*{-1ex}
\paragraph{Spatial-Temporally Lifting.}
Given the absence of ground truth labels, for each image frame of the input video sequences,  we utilize the off-the-shelf 2D vision model to perform panoptic prediction for each pixel $P=(u,v)$, including both instances $\{\hat{R}_t(P)\}_{t=1}^T$ and semantics predictions $\{\hat{O}_t(P)\}_{t=1}^T$. Here, $T$ is the number of time steps (i.e., image frames) to be processed. These predictions are then utilized as the pseudo labels to optimize the panoptic segmentation results from PanoSLAM. However, these predictions usually contain noisy labels that may corrupt the optimization process. 

To address this issue, we unify the panoptic prediction across different time steps over the shared 3D shape through multi-view correspondence. To facilitate this, given two pixels $P^{*1}$ and $P^{*2}$ belonging to different views, if their respective unprojected 3D points $\mathbf{P}^{*1}$ and $\mathbf{P}^{*2}$ are located in the same local small voxel space, we regard these two pixels as being in correspondence.
Specifically, to efficiently identify corresponding pixels across the $T$ frames, we first re-project the pixel depths of each frame to the world coordinates according to their corresponding camera poses and intrinsic (Eq. \ref{equ:project_3D}). These reprojected 3D points are then fused to form a point cloud. The 3D space occupied by this point cloud is uniformly split into voxels with a specified size $S_n$, and we quantize all these points into the centers of voxels directly, allowing us to quickly obtain the corresponding voxel indices. Finally, the points sharing the same voxel index indicate that they are located in the same local small voxel and inherently should have consistent semantic labels for the corresponding pixels across the $T$ frames. This approach allows us to efficiently determine the correspondence of pixels and unify the semantic labels of these pixels.

Specifically, we unprojected the 2D pixel $P_t$ to the 3D location $\mathbf{P}_t \in \mathbb{R}^3$ by following:
\begin{equation}\label{equ:project_3D}
\mathbf{P}_t=E_t^{-1}G^{-1}dP_t,
\end{equation}
where $E_t^{-1}$ is the inverse camera pose at the $t$ time step, and $d$ and $G$ are the ground truth depth and camera intrinsic matrix, respectively. 

In the next, we set all region predictions to be identical within the same local voxel $g_n$, \ie,
\begin{equation}\label{equ:refine}
\hat{R}(P^{*})=\frac{1}{|g_n|}\sum_{* \in g_n}\hat{R}(P^{*}),
\end{equation}
To this end, we obtain the spatial-temporal refined region prediction $\{\hat{R}_t(P^{*})\}_{t=1}^T$. Notably, in our panoptic segmentation formulation Eq. \ref{equ:region_classfication}, the semantics predictions $\{\hat{O}_t(\mathbb{M})\}_{t=1}^T$ of each pixel are adjusted accordingly when their region prediction is changed.

We optimize the Gaussian SLAM system by minimizing a color rendering loss $L_1\left(C_t(P),C_{GT}\right)$, a depth rendering loss $L_1(D_t(P),D_{GT})$, and panoptic prediction loss for $R_t(P)$ and $O_t(\mathbb{M})$. The objective function is as follows:
\begin{equation}\label{equ:loss_function}
\begin{aligned}
\mathbb{L} = \frac{1}{T}\sum_{t \in T}\sum_{P}\lambda_1L_1\left(C_t(P),C_{GT}\right) \\
+\lambda_2L_1(D_t(P),D_{GT}) \\ +\lambda_3CE(O_t(\mathbb{M}),\hat{O}_t(\mathbb{M})) \\
+\lambda_4DICE(R_t(P),\hat{R}_t(P^{*})) \\
+\lambda_5Sig_F(R_t(P),\hat{R}_t(P^{*})),
\end{aligned}
\end{equation}
where $\lambda_{1, ..., 5}$ are the loss weights. $L_1$ indicates the $L_1$ loss, $CE$ is the cross-entropy loss, and $DICE$ and $Sig_F$ are the dice loss and sigmoid focal loss, respectively. Following SplaTAM, we do not optimize over all previous frames, but instead select frames that are most likely to impact the newly added Gaussians. We designate every $u$-th frame as a keyframe and choose $T$ frames for optimization, including the keyframes with the greatest overlap with the current frame. Overlap is determined by analyzing the point cloud of the current frame's depth map and counting the number of points within the frustum of each keyframe.

\begin{table*}[t]
  \centering
  \caption{Quantitative comparison of PanoSLAM with other semantic SLAM methods for semantic segmentation performance on Replica (mIoU(\%). We report the PQ, RQ, and SQ for each scene.}
  \label{tab:semantic_performance}
  \vspace{-1.5ex}
  \scalebox{1}{
  \begin{tabular}{c |c | c |c c c c c c c c}
  \toprule
  w GT& Methods & Metrics &room0 & room1 & room2 & office0& office1& office2& office3& office4 
  \\\midrule
  \multirow{2}{*}{Yes}& NIDS-SLAM & mIoU & 82.45 & 84.08& 76.99& 85.94 &– &– &– &–
  \\ 	
  & DNS SLAM & mIoU& 88.32 &84.90& 81.20& 84.66& – &–& – &– 
  \\
  & SNI-SLAM& mIoU & 88.42 &87.43 &86.16& 87.63& 78.63 &86.49 &74.01 &80.22 
  \\\midrule
  \multirow{8}{*}{No}& \multirow{4}{*}{Baseline} & PQ & 15.2 & 13.3& 12.2 & 5.7 & 16.2 & 11.6 & 14.0 & 16.2
  \\
  &  & SQ & 27.0 & 25.3 & 23.9 & 16.6 & 30.8 & 25.2 & 24.4 & 30.8
  \\
  &  & RQ & 19.5 & 17.1 & 14.7 & 7.1 & \textbf{18.4} & 15.1 & 17.0 & \textbf{18.4}
  \\
  &  & mIoU & 49.07 & 49.80 & 39.96 & 40.06 & \textbf{67.45} & 46.95 &29.04 & 67.45
  \\\cline{2-11}
  & \multirow{4}{*}{PanoSLAM} & PQ & \textbf{19.9} & \textbf{14.1} & \textbf{18.3} & \textbf{10.6} & \textbf{16.7} & \textbf{12.8} & \textbf{14.1} & \textbf{16.3}
  \\
  &  & SQ & \textbf{46.0} & \textbf{26.3} & \textbf{45.9} & \textbf{56.3} & \textbf{33.2} & \textbf{26.3} & \textbf{38.1} & \textbf{41.0}
  \\
  &  & RQ & \textbf{26.6} & \textbf{18.5} & \textbf{21.9} & \textbf{14.2} & 17.6 & \textbf{16.2} & \textbf{18.3} & 13.0
  \\
  &  & mIoU & \textbf{50.32} & \textbf{50.24} & \textbf{44.14} & \textbf{42.34} & 65.2 & \textbf{47.4} & \textbf{37.40} & \textbf{68.3}
  \\\bottomrule
  \end{tabular}}
  \vspace*{-1ex}
\end{table*}

\begin{table*}[t]
  \centering
  \caption{Quantitative comparison of tracking accuracy for our PanoSLAM with other SLAM methods on Replica dataset. We utilize the RMSE (cm) metric as the evaluation metric.}
  \label{tab:tracking}
  \vspace{-1.5ex}
  \scalebox{1}{
  \begin{tabular}{c |c | c c c c c c c c c}
  \toprule
  Type& Methods & room0 & room1 & room2 & office0& office1& office2& office3& office4 & Avg.
  \\\midrule
  \multirow{6}{*}{Visual}& NICE-SLAM & $1.86$ & $2.37$& $2.26$ &$1.50$& $1.01$& $1.85$& $5.67$ &$3.53$ &$2.51$
  \\ 	
  & Co-SLAM & $0.72$ &$0.85$ &$1.02$& $0.69$& $0.56$ &$2.12$ &$1.62$& $0.87$& $1.06$ 
  \\
  & ESLAM & $0.76$& $0.71$ &$0.56$& $0.53$ &$0.49$& $0.58$ &$0.74$& $0.64$ &$0.62$
  \\
  & Point-SLAM & $0.61$& $0.41$ &$0.37$ &\textbf{0.38}& $0.48$& $0.54$ & $0.69$& $0.72$ &$0.52$ 
  \\
  & SplaTAM & \textbf{0.31}& \textbf{0.40} &\textbf{0.29}& $0.47$ &\textbf{0.27} &\textbf{0.29}& \textbf{0.32}& \textbf{0.55} &\textbf{0.36}
  \\\midrule 	
   \multirow{3}{*}{Semantic}& SNI-SLAM & $0.50$ & $0.55$ &$0.45$ &$0.35$ &$0.41$& $0.33$& $0.62$ &\textbf{0.50}& $0.46$ 
  \\
   & DNS SLAM & $0.49$ &$0.46$ &$0.38$ &\textbf{0.34} &$0.35$ &$0.39$ &$0.62$& $0.60$ &$0.45$ 
  \\
   & Ours & \textbf{0.34} & \textbf{0.44} & \textbf{0.24} & 0.48 & \textbf{0.29} & \textbf{0.33} & \textbf{0.52} & 0.52 & \textbf{0.39}
  \\\bottomrule
  \end{tabular}}
  \vspace{-1.5ex}
\end{table*}

\vspace*{-1ex}
\paragraph{SLAM System.}
Our SLAM system is built upon the Gaussian representation and differentiable rendering framework described above. Assuming an existing map represented by a set of 3D Gaussians created from a series of camera frames up to time $t$, when a new RGB-D frame at time $t+1$ is introduced, the SLAM system proceeds through three main steps: Camera Tracking, Gaussian Densification, and Spatial-Temporal Lifting.

First, in the \textbf{Camera Tracking} step, we minimize the image and depth reconstruction errors of the RGB-D frame to optimize the camera pose parameters at time $t+1$, focusing on errors only within the visible silhouette.

Next, the \textbf{Gaussian Densification} step incorporates additional Gaussians into the map based on the rendered silhouette and input depth to enhance the scene representation.

Finally, in the \textbf{Spatial-Temporal Lifting} step, we refine the parameters of all Gaussians in the scene by minimizing RGB, depth and panoptic prediction errors across all images up to frame $t+1$, given the camera poses from frame 1 to $t+1$. This is achieved by optimizing a selected subset of keyframes that overlap with the most recent frame, keeping the batch size manageable for efficient computation.

\section{Experiments}
\label{sec:experiments}

\begin{figure*}
  \centerline{\includegraphics[width=1\textwidth]{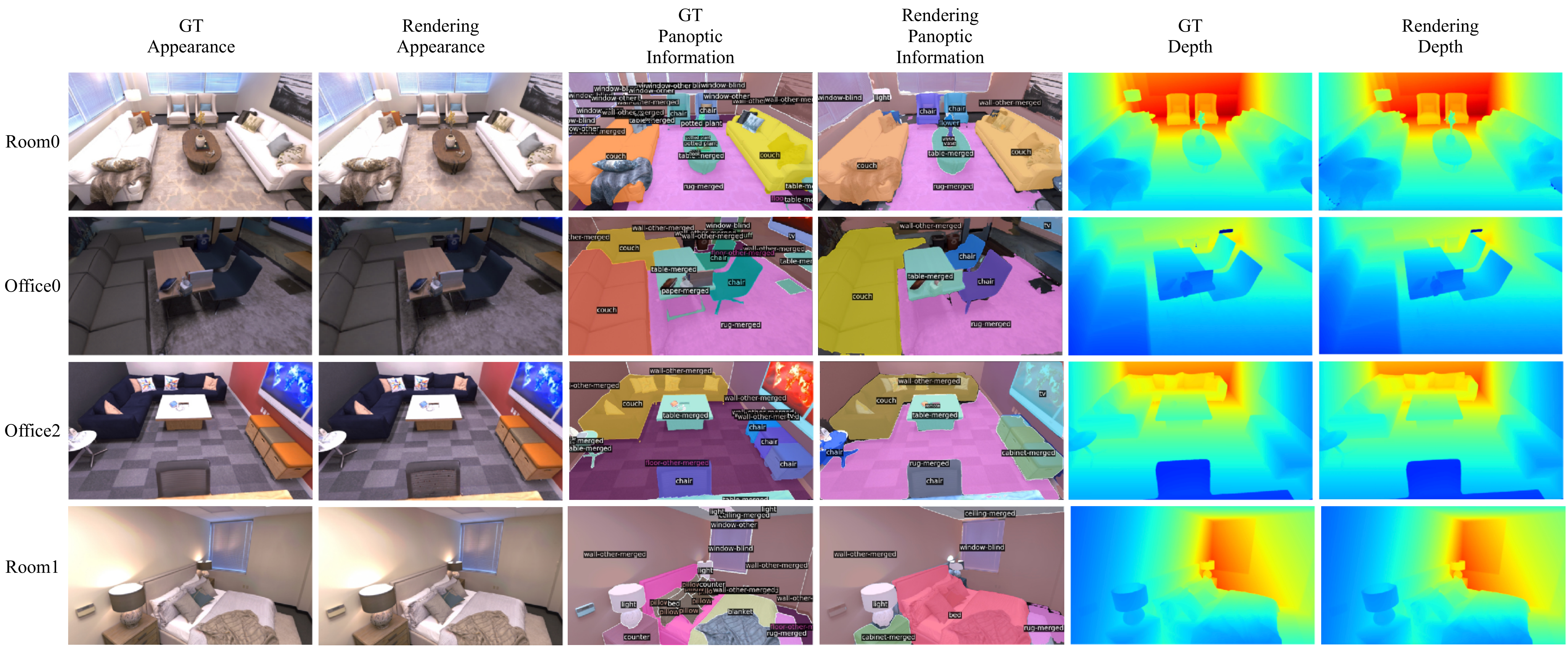}}
  \vspace{-1.5ex}
  \caption{Qualitative results of PanoSLAM in rendering appearance, panoptic information and depth. We present four scenes in the replica dataset, named room0, room1, office0 and office2. Note that different colour in panoptic infomation indicates different instances.}
  \label{fig:visual}
  \vspace{-1ex}
\end{figure*}

\begin{table}[t]
  \centering
  \caption{Evaluation on ScanNet++ (Scene 8b5caf3398 and b20a261fdf) dataset. The experiment result indicates our method's efficiency in real-world scenes.}
  \label{tab:scannetpp}
  \vspace{-1.5ex}
  \scalebox{0.65}{
  \begin{tabular}{c |c |c c c c c c c}
  \toprule
  Scenes&Methods& PQ & SQ & RQ & mIoU & PSNR $\uparrow$ & SSIM$\uparrow$ & LPIPS$\downarrow$
  \\\midrule
  \multirow{2}{*}{8b5caf3398}&base&16.1&26.4&18.3&51.16 &-&-&-
  \\ 	
  &Ours&\textbf{20.3}&\textbf{39.5}&\textbf{33.1}&\textbf{53.26}&\textbf{27.74}&\textbf{0.92}&\textbf{0.13}
  \\\midrule 
  \multirow{2}{*}{b20a261fdf}&base&11.9&18.7&20.4&46.51&-&-&-
  \\ 
  &Ours & \textbf{19.2} & \textbf{42.8}&\textbf{28.3}&\textbf{48.20}&\textbf{28.05}&\textbf{0.91}&\textbf{0.15}
  \\\bottomrule
  \end{tabular}}
  \vspace*{-2ex}
\end{table}

\vspace*{-1ex}
\paragraph{Datasets.}  
We evaluate the performance of PanoSLAM on the Replica \cite{straub2019replica} and ScanNet++ \cite{yeshwanth2023scannet++} datasets, both of which provide ground truth annotations for comprehensive benchmarking. In our experiments, we use eight scenes from the Replica dataset and two scenes from the ScanNet++ dataset to assess PanoSLAM’s mapping, tracking, and segmentation capabilities in diverse environments.

\vspace*{-2ex}
\paragraph{Metrics.}  
To measure the rendering quality and mapping performance, we follow the evaluation metrics defined in \cite{sandstrom2023point}. For mapping accuracy, we use Depth L1 (in cm), and for tracking accuracy, we employ ATE RMSE (in cm). Additionally, RGB image rendering quality is evaluated using PSNR (dB), SSIM, and LPIPS. For semantic segmentation performance, we utilize the widely-used mIoU (mean Intersection-over-Union), which provides a per-pixel accuracy measure consistent with per-pixel classification. For panoptic segmentation, we adopt the standard PQ (Panoptic Quality), SQ (Segmentation Quality), and RQ (Recognition Quality) metrics \cite{kirillov2019panoptic}, reporting single-run results due to the high computational cost associated with training.

\vspace*{-2ex}
\paragraph{Baselines.}  
We compare PanoSLAM with current state-of-the-art dense visual SLAM methods, including NeRF-based SLAM approaches \cite{zhu2022nice,sucar2021imap,johari2023eslam,wang2023co,sandstrom2023point} and 3D Gaussian SLAM, specifically SplaTAM \cite{keetha2023splatam}. For evaluating dense semantic SLAM performance, we benchmark against NeRF-based semantic SLAM methods, including SNI SLAM \cite{zhu2023sni}, DNS SLAM \cite{li2023dns}, and NIDS-SLAM \cite{haghighi2023neural}, which serve as the baselines for comparison in semantic and panoptic segmentation tasks.
\vspace*{-2ex}
\paragraph{Implementation Details.} We employ the SEEM \cite{zou2024segment} 2D vision model to make predictions about the panoptic information. Our framework is built using PyTorch and trained on a RTX 4090 GPU. Throughout the training process, the 2D vision models remain frozen. We conduct warmup training 100 times for the initial five frames in order to initialize the Gaussian semantic rendering. When working with the Replica dataset, the image number set to 4. The loss weights $\lambda_{1,...,5}$ are defined as 1, 1, 1, 1, and 20, respectively. To train the region mask, we utilize the Hungarian algorithm to match the rendering and pseudo masks provided by the vision models.

\begin{table}[t]
  \centering
  \caption{Running time comparison with SplaTam. STL indicates the Spatial-Temporally Lifting. /I and /F indicate running time (ms) per iteration and frame, respectively.}
  \label{tab:Running_time}
  \vspace{-1ex}
  \scalebox{0.7}{
  \begin{tabular}{c |c c c c c c}
  \toprule
  Methods&Tracking/I&Mapping/I&	STL/I&Tracking/F&Mapping/F&STL/F
  \\\midrule
  SplaTam & 21& 22 & -&  890 & 1210 & - 
  \\ 	
  Ours & \textbf{23} & \textbf{24} & \textbf{64} &\textbf{930}&\textbf{1345}&\textbf{642}
  \\\bottomrule
  \end{tabular}}
  \vspace*{-2ex}
\end{table}

\begin{table*}[t]
  \centering
  \caption{Quantitative comparison of training view rendering performance on Replica dataset. Our work outperforms other semantic SLAM methods on all three metrics across all scenes.}
  \label{tab:three_metrics}
  \vspace{-1ex}
  \scalebox{0.78}{
  \begin{tabular}{c |c | c c c c c c c c c c}
  \toprule
  Type & Methods & Metrics & room0 & room1 & room2 & office0& office1& office2& office3& office4 & Avg.
  \\\midrule
  \multirow{12}{*}{Visual}& \multirow{3}{*}{NICE-SLAM}& PSNR$\uparrow$& $22.12$ & $22.47$ &$24.52$ &$29.07$ &$30.34$& $19.66$ &$22.23$& $24.94$& $24.42$
  \\	
  &  & SSIM$\uparrow$ & $0.689$& $0.757$& $0.814$& $0.874$ &$0.886$ &$0.797$& $0.801$& $0.856$ &$0.809$
  \\ 	
  &  & LPIPS$\downarrow$& $0.330$& $0.271$ &$0.208$& $0.229$ &$0.181$ &$0.235$& $0.209$& $0.198$& $0.233$ 
  \\\cline{2-12}
  & \multirow{3}{*}{Co-SLAM}& PSNR$\uparrow$& $27.27$ &$28.45$& $29.06$ &$34.14$ &$34.87$ &$28.43$& $28.76$& $30.91$ &$30.24$
  \\	
  &  & SSIM$\uparrow$& $0.910$& $0.909$& $0.932$ &$0.961$& $0.969$ &$0.938$ &$0.941$& $0.955$ &$0.939$
  \\ 	
  & & LPIPS$\downarrow$& $0.324$& $0.294$& $0.266$ &$0.209$ &$0.196$ &$0.258$& $0.229$ &$0.236$ &$0.252$ 
  \\\cline{2-12}
  & \multirow{3}{*}{ESLAM}& PSNR$\uparrow$& $25.32$& $27.77$& $29.08$ &$33.71$ &$30.20$& $28.09$ &$28.77$ &$29.71$ &$29.08$
  \\	
  &  & SSIM$\uparrow$& $0.875$& $0.902$& $0.932$& $0.960$ &$0.923$ &$0.943$ &$0.948$ &$0.945$ &$0.929$
  \\ 	
  & & LPIPS$\downarrow$& $0.313$ &$0.298$& $0.248$ &$0.184$& $0.228$ &$0.241$& $0.196$ &$0.204$ &$0.239$ 
  \\\cline{2-12}
  & \multirow{3}{*}{SplaTAM}& PSNR$\uparrow$& $32.86$& $33.89$ &$35.25$ &$38.26$& $39.17$& $31.97$ &$29.70$ &$31.81$& $34.11$
  \\	
  &  & SSIM$\uparrow$& $0.978$ &$0.969$ &$0.979$ &$0.977$ &$0.978$ &$0.968$ &$0.949$ &$0.949$& $0.968$
  \\ 	
  & & LPIPS$\downarrow$& $0.072$& $0.103$& $0.081$ &$0.092$ &$0.093$ &$0.102$ &$0.121$ &$0.152$& $0.102$ 
  \\\midrule
  \multirow{6}{*}{Semantic}& \multirow{3}{*}{SNI-SLAM}& PSNR$\uparrow$& $25.91$& $28.17$& $29.15$ &$33.86$ &$30.34$ &$29.10$ &$29.02$ &$29.87$& $29.43$
  \\	
  &  & SSIM$\uparrow$& $0.885$ &$0.910$& $0.938$ &$0.965$& $0.927$& $0.950$ &\textbf{0.950} &$0.952$ &$0.935$
  \\ 	
  & & LPIPS$\downarrow$& $0.307$&$0.292$ &$0.245$ &$0.182$ &$0.225$& $0.238$& $0.192$ &$0.198$ &$0.235$ 
  \\\cline{2-12}
  & \multirow{3}{*}{Ours}& PSNR$\uparrow$& \textbf{32.89} & \textbf{33.05} & \textbf{34.24} & \textbf{37.90} & \textbf{38.29} & \textbf{29.61} & \textbf{29.82} & \textbf{31.02} & \textbf{33.35}
  \\	
  &  & SSIM$\uparrow$& \textbf{0.979} & \textbf{0.960} & \textbf{0.982} & \textbf{0.980} & \textbf{0.980} & \textbf{0.952} & 0.943 & \textbf{0.939} & \textbf{0.964}
  \\ 	
  & & LPIPS$\downarrow$& \textbf{0.071} & \textbf{0.115} & \textbf{0.082} & \textbf{0.092} & \textbf{0.095} & \textbf{0.121} & \textbf{0.127} & \textbf{0.166} & \textbf{0.108}
  \\\bottomrule
  \end{tabular}}
\end{table*}

\begin{table*}[!ht]
  \centering
  \caption{Comparion of our method with existing radiance field-based SLAM methods on Replica for reconstruction metric Depth L1 (cm).}
  \label{tab:Depth}
  \vspace{-1ex}
  \scalebox{0.85}{
  \begin{tabular}{c |c | c c c c c c c c c}
  \toprule
  Type& Methods & room0 & room1 & room2 & office0& office1& office2& office3& office4 & Avg.
  \\\midrule
  \multirow{4}{*}{Visual}& NICE-SLAM & 1.81 & 1.44 &2.04& 1.39 &1.76 &8.33 &4.99 &2.01& 2.97
  \\ 	
  & Vox-Fusion& 1.09 &1.90 &2.21 &2.32& 3.40& 4.19& 2.96& 1.61 &2.46 
  \\
  & Co-SLAM & 1.05 &0.85& 2.37& 1.24 &1.48 &1.86 &1.66& 1.54 &1.51 
  \\
  & ESLAM & 0.73 &0.74& 1.26 &0.71 &1.02& 0.93 &1.03 &1.18& 0.95
  \\\midrule 
  \multirow{2}{*}{Semantic}& SNI-SLAM & 0.55 &0.58 &0.87& 0.55 &0.97 &0.89 &0.75& \textbf{0.97}& 0.77 
  \\	
   & Ours & \textbf{0.52} & \textbf{0.51} & \textbf{0.53} & \textbf{0.38} & \textbf{0.29} & \textbf{0.85} & \textbf{0.70} & 1.15 & \textbf{0.61}
  \\\bottomrule
  \end{tabular}}
\end{table*}

\begin{table}[t]
  \centering
  \caption{Ablation experiments on room0. We report the panoptic segmentation performance.}
  \label{tab:ablation}
  \vspace{-1ex}
  \scalebox{0.8}{
  \begin{tabular}{c |c c c c c c c}
  \toprule
  Setting& PQ & SQ & RQ & mIoU & PSNR $\uparrow$ & SSIM$\uparrow$ & LPIPS$\downarrow$
  \\\midrule
  base & 15.2& 27.0 & 19.5 & 49.07 &- &-&-
  \\ 	
  w/o STL& 7.3& 18.2 & 10.3 & 40.05 &29.29 &0.918&0.113
  \\ 	
  STL-2& 11.6& 36.1 & 17.1 & 46.87 &30.31 &0.945&0.101
  \\ 
  Ours & \textbf{19.9} & \textbf{46.0} & \textbf{26.6} &\textbf{50.32}&\textbf{32.89}&\textbf{0.979}&\textbf{0.071}
  \\\bottomrule
  \end{tabular}}
\end{table}

\subsection{Results and Discussion}
\vspace{-1ex}
We performed a comprehensive quantitative comparison of our method against state-of-the-art approaches across multiple dimensions, including segmentation (Tab. \ref{tab:semantic_performance}), tracking (Tab. \ref{tab:tracking}), rendering (Tab. \ref{tab:three_metrics}), and reconstruction (Tab. \ref{tab:Depth}) results. Besides, we present qualitative results in Fig. \ref{fig:visual}.

\vspace*{-2ex}
\paragraph{Panoptic and Semantic Segmentation.}  
The panoptic and semantic segmentation results are shown in Tables \ref{tab:semantic_performance} and \ref{tab:scannetpp}. As the first label-free semantic SLAM, we compare our method with the baseline SEEM’s predictions. Our method demonstrates improved performance on both the Replica and ScanNet++ datasets, highlighting its effectiveness in label-free panoptic and semantic segmentation.

\vspace*{-2ex}
\paragraph{Tracking.}  
As shown in Tab. \ref{tab:tracking}, our approach achieves superior tracking accuracy compared to other semantic SLAM methods. This improvement is largely due to the Spatial-Temporal Lifting technique, which ensures consistent semantic information over time and across viewpoints, effectively reducing accumulated drift in the tracking process.

\vspace*{-2ex}
\paragraph{Rendering.}  
Table \ref{tab:three_metrics} presents the rendering quality on input views from the Replica dataset. Our method achieves the highest performance across PSNR, SSIM, and LPIPS metrics compared to other dense semantic SLAM methods, indicating superior visual fidelity.

\vspace*{-2ex}
\paragraph{Reconstruction.}  
As shown in Tab. \ref{tab:Depth}, our method outperforms other semantic SLAM methods in reconstruction accuracy, demonstrating its superior effectiveness and efficiency in the 3D mapping process.

\vspace*{-2ex}
\paragraph{Qualitative Results.}  
Fig. \ref{fig:visual} shows qualitative evaluations of PanoSLAM, highlighting its performance in rendering appearance, semantics, and depth. Notably, PanoSLAM achieves impressive panoptic and semantic segmentation results without any manual labels, underscoring its robustness in label-free environments.

\vspace*{-2ex}
\paragraph{Running Time Analysis.}  
Table \ref{tab:Running_time} provides a running time analysis, showing that our method incurs slightly higher per-iteration times for tracking (19.9 ms vs. 15.2 ms) and mapping (46.0 ms vs. 27.0 ms) compared to SplaTAM, due to the added panoptic processing. Nevertheless, the per-frame times remain efficient, with minimal overhead from the STL module (0.979 ms), confirming the overall efficiency of our approach.

\vspace*{-1ex}
\subsection{Ablation Study}  
We performed a series of ablation experiments on the room0 scene to evaluate the effectiveness of various components in our framework. As shown in Tab. \ref{tab:ablation}, "Base" represents the predictions from the 2D vision model alone, while "w/o STL" indicates that the Spatial-Temporal Lifting (STL) module was not used during training. "STL-(n)" represents the inclusion of STL with \(n\) time-steps of images, with \(n\) set to 4 in our full method. The results of these ablation studies highlight the critical role of STL in enhancing the ability of 3D Gaussians to effectively learn from the noisy pseudo-labels generated by the 2D vision model. This emphasizes the significant impact of STL within our framework.

\section{Conclusions}
\label{sec:conclusions}
We introduce PanoSLAM, the first Gaussian-based SLAM method capable of reconstructing panoptic 3D scene from unlabeled RGB-D videos. To effectively distill knowledge from 2D vision foundation models into a 3D Gaussian splatting SLAM framework, we propose a novel Spatial-Temporal Lifting module. Experimental results demonstrate that our method significantly outperforms state-of-the-art approaches. Furthermore, for the first time, we successfully recover panoptic information in 3D open-world scenes without any manual labels.

\vspace*{-2ex}
\paragraph{Limitations and Future Work.}
Currently, our approach relies on 2D vision foundation models to generate pseudo-labels for guiding semantic reconstruction. However, these labels can be noisy, particularly in areas with fine and intricate details, such as flower leaves in large, complex rooms. Despite the improvements provided by our Spatial-Temporal Lifting module, achieving precise semantic reconstruction in these regions remains challenging. In future work, we aim to explore ways to integrate multi-view information into 2D vision foundation models to produce more accurate and detailed semantic labels, ultimately enhancing the quality of semantic scene reconstruction.

{
    \small
    \bibliographystyle{ieeenat_fullname}
    \bibliography{main}

\begin{thebibliography}{73}
\providecommand{\natexlab}[1]{#1}
\providecommand{\url}[1]{\texttt{#1}}
\expandafter\ifx\csname urlstyle\endcsname\relax
  \providecommand{\doi}[1]{doi: #1}\else
  \providecommand{\doi}{doi: \begingroup \urlstyle{rm}\Url}\fi

\bibitem[Bucher et~al.(2019)Bucher, Vu, Cord, and P{\'e}rez]{bucher2019zero}
Maxime Bucher, Tuan-Hung Vu, Matthieu Cord, and Patrick P{\'e}rez.
\newblock Zero-shot semantic segmentation.
\newblock \emph{Advances in Neural Information Processing Systems}, 32, 2019.

\bibitem[Chen et~al.(2022)Chen, Zhu, Chen, Li, Ma, Yang, and Wang]{chen2022zero}
Runnan Chen, Xinge Zhu, Nenglun Chen, Wei Li, Yuexin Ma, Ruigang Yang, and Wenping Wang.
\newblock Zero-shot point cloud segmentation by transferring geometric primitives.
\newblock \emph{arXiv preprint arXiv:2210.09923}, 2022.

\bibitem[Chen et~al.(2023{\natexlab{a}})Chen, Liu, Kong, Zhu, Ma, Li, Hou, Qiao, and Wang]{chen2023clip2scene}
Runnan Chen, Youquan Liu, Lingdong Kong, Xinge Zhu, Yuexin Ma, Yikang Li, Yuenan Hou, Yu Qiao, and Wenping Wang.
\newblock Clip2scene: Towards label-efficient 3d scene understanding by clip.
\newblock In \emph{Proceedings of the IEEE/CVF Conference on Computer Vision and Pattern Recognition}, pages 7020--7030, 2023{\natexlab{a}}.

\bibitem[Chen et~al.(2023{\natexlab{b}})Chen, Zhu, Chen, Li, Ma, Yang, and Wang]{chen2023bridging}
Runnan Chen, Xinge Zhu, Nenglun Chen, Wei Li, Yuexin Ma, Ruigang Yang, and Wenping Wang.
\newblock Bridging language and geometric primitives for zero-shot point cloud segmentation.
\newblock In \emph{Proceedings of the 31st ACM International Conference on Multimedia}, pages 5380--5388, 2023{\natexlab{b}}.

\bibitem[Chen et~al.(2023{\natexlab{c}})Chen, Zhu, Chen, Wang, Li, Ma, Yang, Liu, and Wang]{chen2023model2scene}
Runnan Chen, Xinge Zhu, Nenglun Chen, Dawei Wang, Wei Li, Yuexin Ma, Ruigang Yang, Tongliang Liu, and Wenping Wang.
\newblock Model2scene: Learning 3d scene representation via contrastive language-cad models pre-training.
\newblock \emph{arXiv preprint arXiv:2309.16956}, 2023{\natexlab{c}}.

\bibitem[Chen et~al.(2024)Chen, Liu, Kong, Chen, Zhu, Ma, Liu, and Wang]{chen2024towards}
Runnan Chen, Youquan Liu, Lingdong Kong, Nenglun Chen, Xinge Zhu, Yuexin Ma, Tongliang Liu, and Wenping Wang.
\newblock Towards label-free scene understanding by vision foundation models.
\newblock \emph{Advances in Neural Information Processing Systems}, 36, 2024.

\bibitem[Cheng et~al.(2020)Cheng, Collins, Zhu, Liu, Huang, Adam, and Chen]{cheng2020panoptic}
Bowen Cheng, Maxwell~D Collins, Yukun Zhu, Ting Liu, Thomas~S Huang, Hartwig Adam, and Liang-Chieh Chen.
\newblock Panoptic-deeplab: A simple, strong, and fast baseline for bottom-up panoptic segmentation.
\newblock In \emph{Proceedings of the IEEE/CVF conference on computer vision and pattern recognition}, pages 12475--12485, 2020.

\bibitem[Cheng et~al.(2021{\natexlab{a}})Cheng, Schwing, and Kirillov]{cheng2021per}
Bowen Cheng, Alex Schwing, and Alexander Kirillov.
\newblock Per-pixel classification is not all you need for semantic segmentation.
\newblock \emph{Advances in neural information processing systems}, 34:\penalty0 17864--17875, 2021{\natexlab{a}}.

\bibitem[Cheng et~al.(2022)Cheng, Misra, Schwing, Kirillov, and Girdhar]{cheng2022masked}
Bowen Cheng, Ishan Misra, Alexander~G Schwing, Alexander Kirillov, and Rohit Girdhar.
\newblock Masked-attention mask transformer for universal image segmentation.
\newblock In \emph{Proceedings of the IEEE/CVF Conference on Computer Vision and Pattern Recognition}, pages 1290--1299, 2022.

\bibitem[Cheng et~al.(2021{\natexlab{b}})Cheng, Razani, Taghavi, Li, and Liu]{af2s3net}
Ran Cheng, Ryan Razani, Ehsan Taghavi, Enxu Li, and Bingbing Liu.
\newblock (af)2-s3net: Attentive feature fusion with adaptive feature selection for sparse semantic segmentation network.
\newblock In \emph{IEEE Conference on Computer Vision and Pattern Recognition}, pages 12547--12556, 2021{\natexlab{b}}.

\bibitem[Contributors(2020)]{contributors2020mmdetection3d}
MMDetection3D Contributors.
\newblock Mmdetection3d: Openmmlab next-generation platform for general 3d object detection, 2020.

\bibitem[Davison et~al.(2007)Davison, Reid, Molton, and Stasse]{davison2007monoslam}
Andrew~J Davison, Ian~D Reid, Nicholas~D Molton, and Olivier Stasse.
\newblock Monoslam: Real-time single camera slam.
\newblock \emph{IEEE transactions on pattern analysis and machine intelligence}, 29\penalty0 (6):\penalty0 1052--1067, 2007.

\bibitem[Ding et~al.(2022)Ding, Yang, Xue, Zhang, Bai, and Qi]{ding2022language}
Runyu Ding, Jihan Yang, Chuhui Xue, Wenqing Zhang, Song Bai, and Xiaojuan Qi.
\newblock Language-driven open-vocabulary 3d scene understanding.
\newblock \emph{arXiv preprint arXiv:2211.16312}, 2022.

\bibitem[Haghighi et~al.(2023)Haghighi, Kumar, Thiran, and Van~Gool]{haghighi2023neural}
Yasaman Haghighi, Suryansh Kumar, Jean-Philippe Thiran, and Luc Van~Gool.
\newblock Neural implicit dense semantic slam.
\newblock \emph{arXiv preprint arXiv:2304.14560}, 2023.

\bibitem[Hong et~al.(2022)Hong, Kong, Zhou, Zhu, Li, and Liu]{hong2022dsnet}
Fangzhou Hong, Lingdong Kong, Hui Zhou, Xinge Zhu, Hongsheng Li, and Ziwei Liu.
\newblock Unified 3d and 4d panoptic segmentation via dynamic shifting network.
\newblock \emph{arXiv preprint arXiv:2203.07186}, 2022.

\bibitem[Hu et~al.(2020)Hu, Sclaroff, and Saenko]{hu2020uncertainty}
Ping Hu, Stan Sclaroff, and Kate Saenko.
\newblock Uncertainty-aware learning for zero-shot semantic segmentation.
\newblock \emph{Advances in Neural Information Processing Systems}, 33:\penalty0 21713--21724, 2020.

\bibitem[Johari et~al.(2023)Johari, Carta, and Fleuret]{johari2023eslam}
Mohammad~Mahdi Johari, Camilla Carta, and Francois Fleuret.
\newblock Eslam: Efficient dense slam system based on hybrid representation of signed distance fields.
\newblock In \emph{Proceedings of the IEEE/CVF Conference on Computer Vision and Pattern Recognition}, pages 17408--17419, 2023.

\bibitem[Keetha et~al.(2023)Keetha, Karhade, Jatavallabhula, Yang, Scherer, Ramanan, and Luiten]{keetha2023splatam}
Nikhil Keetha, Jay Karhade, Krishna~Murthy Jatavallabhula, Gengshan Yang, Sebastian Scherer, Deva Ramanan, and Jonathon Luiten.
\newblock Splatam: Splat, track \& map 3d gaussians for dense rgb-d slam.
\newblock \emph{arXiv preprint arXiv:2312.02126}, 2023.

\bibitem[Kerbl et~al.(2023)Kerbl, Kopanas, Leimk{\"u}hler, and Drettakis]{kerbl20233d}
Bernhard Kerbl, Georgios Kopanas, Thomas Leimk{\"u}hler, and George Drettakis.
\newblock 3d gaussian splatting for real-time radiance field rendering.
\newblock \emph{ACM Transactions on Graphics}, 42\penalty0 (4):\penalty0 1--14, 2023.

\bibitem[Kerr et~al.(2023)Kerr, Kim, Goldberg, Kanazawa, and Tancik]{kerr2023lerf}
Justin Kerr, Chung~Min Kim, Ken Goldberg, Angjoo Kanazawa, and Matthew Tancik.
\newblock Lerf: Language embedded radiance fields.
\newblock In \emph{Proceedings of the IEEE/CVF International Conference on Computer Vision}, pages 19729--19739, 2023.

\bibitem[Keselman and Hebert(2022)]{keselman2022approximate}
Leonid Keselman and Martial Hebert.
\newblock Approximate differentiable rendering with algebraic surfaces.
\newblock In \emph{European Conference on Computer Vision}, pages 596--614. Springer, 2022.

\bibitem[Keselman and Hebert(2023)]{keselman2023flexible}
Leonid Keselman and Martial Hebert.
\newblock Flexible techniques for differentiable rendering with 3d gaussians.
\newblock \emph{arXiv preprint arXiv:2308.14737}, 2023.

\bibitem[Kirillov et~al.(2019)Kirillov, He, Girshick, Rother, and Doll{\'a}r]{kirillov2019panoptic}
Alexander Kirillov, Kaiming He, Ross Girshick, Carsten Rother, and Piotr Doll{\'a}r.
\newblock Panoptic segmentation.
\newblock In \emph{Proceedings of the IEEE/CVF conference on computer vision and pattern recognition}, pages 9404--9413, 2019.

\bibitem[Kirillov et~al.(2023)Kirillov, Mintun, Ravi, Mao, Rolland, Gustafson, Xiao, Whitehead, Berg, Lo, et~al.]{kirillov2023segment}
Alexander Kirillov, Eric Mintun, Nikhila Ravi, Hanzi Mao, Chloe Rolland, Laura Gustafson, Tete Xiao, Spencer Whitehead, Alexander~C Berg, Wan-Yen Lo, et~al.
\newblock Segment anything.
\newblock In \emph{Proceedings of the IEEE/CVF International Conference on Computer Vision}, pages 4015--4026, 2023.

\bibitem[Klein and Murray(2007)]{klein2007parallel}
Georg Klein and David Murray.
\newblock Parallel tracking and mapping for small ar workspaces.
\newblock In \emph{2007 6th IEEE and ACM international symposium on mixed and augmented reality}, pages 225--234. IEEE, 2007.

\bibitem[Kong et~al.(2023{\natexlab{a}})Kong, Liu, Chen, Ma, Zhu, Li, Hou, Qiao, and Liu]{kong2023rethinking}
Lingdong Kong, Youquan Liu, Runnan Chen, Yuexin Ma, Xinge Zhu, Yikang Li, Yuenan Hou, Yu Qiao, and Ziwei Liu.
\newblock Rethinking range view representation for lidar segmentation.
\newblock In \emph{Proceedings of the IEEE/CVF International Conference on Computer Vision}, pages 228--240, 2023{\natexlab{a}}.

\bibitem[Kong et~al.(2023{\natexlab{b}})Kong, Liu, Li, Chen, Zhang, Ren, Pan, Chen, and Liu]{kong2023robo3d}
Lingdong Kong, Youquan Liu, Xin Li, Runnan Chen, Wenwei Zhang, Jiawei Ren, Liang Pan, Kai Chen, and Ziwei Liu.
\newblock Robo3d: Towards robust and reliable 3d perception against corruptions.
\newblock In \emph{Proceedings of the IEEE/CVF International Conference on Computer Vision}, pages 19994--20006, 2023{\natexlab{b}}.

\bibitem[Li et~al.(2022)Li, Weinberger, Belongie, Koltun, and Ranftl]{lilanguage}
Boyi Li, Kilian~Q Weinberger, Serge Belongie, Vladlen Koltun, and Rene Ranftl.
\newblock Language-driven semantic segmentation.
\newblock In \emph{International Conference on Learning Representations}, 2022.

\bibitem[Li et~al.(2023)Li, Niemeyer, Navab, and Tombari]{li2023dns}
Kunyi Li, Michael Niemeyer, Nassir Navab, and Federico Tombari.
\newblock Dns slam: Dense neural semantic-informed slam.
\newblock \emph{arXiv preprint arXiv:2312.00204}, 2023.

\bibitem[Li et~al.(2024{\natexlab{a}})Li, Liu, and Zhou]{li2024sgs}
Mingrui Li, Shuhong Liu, and Heng Zhou.
\newblock Sgs-slam: Semantic gaussian splatting for neural dense slam.
\newblock \emph{arXiv preprint arXiv:2402.03246}, 2024{\natexlab{a}}.

\bibitem[Li et~al.(2020)Li, Wei, and Yang]{li2020consistent}
Peike Li, Yunchao Wei, and Yi Yang.
\newblock Consistent structural relation learning for zero-shot segmentation.
\newblock \emph{Advances in Neural Information Processing Systems}, 33:\penalty0 10317--10327, 2020.

\bibitem[Li et~al.(2024{\natexlab{b}})Li, Huang, Chen, Che, Guo, Liu, Karray, and Gong]{li2024urban4d}
Ziwen Li, Jiaxin Huang, Runnan Chen, Yunlong Che, Yandong Guo, Tongliang Liu, Fakhri Karray, and Mingming Gong.
\newblock Urban4d: Semantic-guided 4d gaussian splatting for urban scene reconstruction.
\newblock \emph{arXiv preprint arXiv:2412.03473}, 2024{\natexlab{b}}.

\bibitem[Liu et~al.(2023)Liu, Chen, Li, Kong, Yang, Xia, Bai, Zhu, Ma, Li, et~al.]{liu2023uniseg}
Youquan Liu, Runnan Chen, Xin Li, Lingdong Kong, Yuchen Yang, Zhaoyang Xia, Yeqi Bai, Xinge Zhu, Yuexin Ma, Yikang Li, et~al.
\newblock Uniseg: A unified multi-modal lidar segmentation network and the openpcseg codebase.
\newblock In \emph{Proceedings of the IEEE/CVF International Conference on Computer Vision}, pages 21662--21673, 2023.

\bibitem[Liu et~al.(2024)Liu, Kong, Wu, Chen, Li, Pan, Liu, and Ma]{liu2024multi}
Youquan Liu, Lingdong Kong, Xiaoyang Wu, Runnan Chen, Xin Li, Liang Pan, Ziwei Liu, and Yuexin Ma.
\newblock Multi-space alignments towards universal lidar segmentation.
\newblock In \emph{Proceedings of the IEEE/CVF Conference on Computer Vision and Pattern Recognition}, pages 14648--14661, 2024.

\bibitem[Lu et~al.(2023)Lu, Jiang, Chen, Hou, Zhu, and Ma]{lu2023see}
Yuhang Lu, Qi Jiang, Runnan Chen, Yuenan Hou, Xinge Zhu, and Yuexin Ma.
\newblock See more and know more: Zero-shot point cloud segmentation via multi-modal visual data.
\newblock In \emph{Proceedings of the IEEE/CVF International Conference on Computer Vision}, pages 21674--21684, 2023.

\bibitem[Matsuki et~al.(2023)Matsuki, Murai, Kelly, and Davison]{matsuki2023gaussian}
Hidenobu Matsuki, Riku Murai, Paul~HJ Kelly, and Andrew~J Davison.
\newblock Gaussian splatting slam.
\newblock \emph{arXiv preprint arXiv:2312.06741}, 2023.

\bibitem[Michele et~al.(2021)Michele, Boulch, Puy, Bucher, and Marlet]{michele2021generative}
Bj{\"o}rn Michele, Alexandre Boulch, Gilles Puy, Maxime Bucher, and Renaud Marlet.
\newblock Generative zero-shot learning for semantic segmentation of 3d point clouds.
\newblock In \emph{International Conference on 3D Vision}, pages 992--1002, 2021.

\bibitem[Mur-Artal et~al.(2015)Mur-Artal, Montiel, and Tardos]{mur2015orb}
Raul Mur-Artal, Jose Maria~Martinez Montiel, and Juan~D Tardos.
\newblock Orb-slam: a versatile and accurate monocular slam system.
\newblock \emph{IEEE transactions on robotics}, 31\penalty0 (5):\penalty0 1147--1163, 2015.

\bibitem[Peng et~al.(2022)Peng, Genova, Jiang, Tagliasacchi, Pollefeys, Funkhouser, et~al.]{peng2022openscene}
Songyou Peng, Kyle Genova, Chiyu Jiang, Andrea Tagliasacchi, Marc Pollefeys, Thomas Funkhouser, et~al.
\newblock Openscene: 3d scene understanding with open vocabularies.
\newblock \emph{arXiv preprint arXiv:2211.15654}, 2022.

\bibitem[Peng et~al.(2023)Peng, Genova, Jiang, Tagliasacchi, Pollefeys, Funkhouser, et~al.]{peng2023openscene}
Songyou Peng, Kyle Genova, Chiyu Jiang, Andrea Tagliasacchi, Marc Pollefeys, Thomas Funkhouser, et~al.
\newblock Openscene: 3d scene understanding with open vocabularies.
\newblock In \emph{Proceedings of the IEEE/CVF conference on computer vision and pattern recognition}, pages 815--824, 2023.

\bibitem[Peng et~al.(2025)Peng, Chen, Qiao, Kong, Liu, Sun, Wang, Zhu, and Ma]{peng2025learning}
Xidong Peng, Runnan Chen, Feng Qiao, Lingdong Kong, Youquan Liu, Yujing Sun, Tai Wang, Xinge Zhu, and Yuexin Ma.
\newblock Learning to adapt sam for segmenting cross-domain point clouds.
\newblock In \emph{European Conference on Computer Vision}, pages 54--71. Springer, 2025.

\bibitem[Qi et~al.(2017)Qi, Su, Mo, and Guibas]{qi2017pointnet}
Charles~R Qi, Hao Su, Kaichun Mo, and Leonidas~J Guibas.
\newblock Pointnet: Deep learning on point sets for 3d classification and segmentation.
\newblock In \emph{IEEE/CVF Conference on Computer Vision and Pattern Recognition}, pages 652--660, 2017.

\bibitem[Qin et~al.(2024)Qin, Li, Zhou, Wang, and Pfister]{qin2024langsplat}
Minghan Qin, Wanhua Li, Jiawei Zhou, Haoqian Wang, and Hanspeter Pfister.
\newblock Langsplat: 3d language gaussian splatting.
\newblock In \emph{Proceedings of the IEEE/CVF Conference on Computer Vision and Pattern Recognition}, pages 20051--20060, 2024.

\bibitem[Radford et~al.(2021)Radford, Kim, Hallacy, Ramesh, Goh, Agarwal, Sastry, Askell, Mishkin, Clark, et~al.]{radford2021learning}
Alec Radford, Jong~Wook Kim, Chris Hallacy, Aditya Ramesh, Gabriel Goh, Sandhini Agarwal, Girish Sastry, Amanda Askell, Pamela Mishkin, Jack Clark, et~al.
\newblock Learning transferable visual models from natural language supervision.
\newblock In \emph{International Conference on Machine Learning}, pages 8748--8763. PMLR, 2021.

\bibitem[Riz et~al.(2023)Riz, Saltori, Ricci, and Poiesi]{riz2023novel}
Luigi Riz, Cristiano Saltori, Elisa Ricci, and Fabio Poiesi.
\newblock Novel class discovery for 3d point cloud semantic segmentation.
\newblock \emph{arXiv preprint arXiv:2303.11610}, 2023.

\bibitem[Rosinol et~al.(2020)Rosinol, Abate, Chang, and Carlone]{rosinol2020kimera}
Antoni Rosinol, Marcus Abate, Yun Chang, and Luca Carlone.
\newblock Kimera: an open-source library for real-time metric-semantic localization and mapping.
\newblock In \emph{IEEE International Conference on Robotics and Automation}, pages 1689--1696. IEEE, 2020.

\bibitem[Salas-Moreno et~al.(2013)Salas-Moreno, Newcombe, Strasdat, Kelly, and Davison]{salas2013slam}
Renato~F Salas-Moreno, Richard~A Newcombe, Hauke Strasdat, Paul~HJ Kelly, and Andrew~J Davison.
\newblock Slam++: Simultaneous localisation and mapping at the level of objects.
\newblock In \emph{Proceedings of the IEEE conference on computer vision and pattern recognition}, pages 1352--1359, 2013.

\bibitem[Sandstr{\"o}m et~al.(2023)Sandstr{\"o}m, Li, Van~Gool, and Oswald]{sandstrom2023point}
Erik Sandstr{\"o}m, Yue Li, Luc Van~Gool, and Martin~R Oswald.
\newblock Point-slam: Dense neural point cloud-based slam.
\newblock In \emph{Proceedings of the IEEE/CVF International Conference on Computer Vision}, pages 18433--18444, 2023.

\bibitem[Sautier et~al.(2022)Sautier, Puy, Gidaris, Boulch, Bursuc, and Marlet]{sautier2022image}
Corentin Sautier, Gilles Puy, Spyros Gidaris, Alexandre Boulch, Andrei Bursuc, and Renaud Marlet.
\newblock Image-to-lidar self-supervised distillation for autonomous driving data.
\newblock In \emph{Proceedings of the IEEE/CVF Conference on Computer Vision and Pattern Recognition}, pages 9891--9901, 2022.

\bibitem[Shi et~al.(2023)Shi, Wang, Duan, and Guan]{shi2023language}
Jin-Chuan Shi, Miao Wang, Hao-Bin Duan, and Shao-Hua Guan.
\newblock Language embedded 3d gaussians for open-vocabulary scene understanding.
\newblock \emph{arXiv preprint arXiv:2311.18482}, 2023.

\bibitem[Siddiqui et~al.(2023)Siddiqui, Porzi, Bul{\'o}, M{\"u}ller, Nie{\ss}ner, Dai, and Kontschieder]{siddiqui2023panoptic}
Yawar Siddiqui, Lorenzo Porzi, Samuel~Rota Bul{\'o}, Norman M{\"u}ller, Matthias Nie{\ss}ner, Angela Dai, and Peter Kontschieder.
\newblock Panoptic lifting for 3d scene understanding with neural fields.
\newblock In \emph{Proceedings of the IEEE/CVF Conference on Computer Vision and Pattern Recognition}, pages 9043--9052, 2023.

\bibitem[Straub et~al.(2019)Straub, Whelan, Ma, Chen, Wijmans, Green, Engel, Mur-Artal, Ren, Verma, et~al.]{straub2019replica}
Julian Straub, Thomas Whelan, Lingni Ma, Yufan Chen, Erik Wijmans, Simon Green, Jakob~J Engel, Raul Mur-Artal, Carl Ren, Shobhit Verma, et~al.
\newblock The replica dataset: A digital replica of indoor spaces.
\newblock \emph{arXiv preprint arXiv:1906.05797}, 2019.

\bibitem[Strudel et~al.(2021)Strudel, Garcia, Laptev, and Schmid]{strudel2021segmenter}
Robin Strudel, Ricardo Garcia, Ivan Laptev, and Cordelia Schmid.
\newblock Segmenter: Transformer for semantic segmentation.
\newblock In \emph{Proceedings of the IEEE/CVF international conference on computer vision}, pages 7262--7272, 2021.

\bibitem[Sucar et~al.(2021)Sucar, Liu, Ortiz, and Davison]{sucar2021imap}
Edgar Sucar, Shikun Liu, Joseph Ortiz, and Andrew~J Davison.
\newblock imap: Implicit mapping and positioning in real-time.
\newblock In \emph{Proceedings of the IEEE/CVF International Conference on Computer Vision}, pages 6229--6238, 2021.

\bibitem[Sun et~al.(2024)Sun, Qing, Xu, Kong, Liu, Li, Zhu, Zhang, Xiao, Chen, et~al.]{sun2024empirical}
Jiahao Sun, Chunmei Qing, Xiang Xu, Lingdong Kong, Youquan Liu, Li Li, Chenming Zhu, Jingwei Zhang, Zeqi Xiao, Runnan Chen, et~al.
\newblock An empirical study of training state-of-the-art lidar segmentation models.
\newblock \emph{arXiv preprint arXiv:2405.14870}, 2024.

\bibitem[Wang et~al.(2022)Wang, Wang, Sun, Kortylewski, and Yuille]{wang2022voge}
Angtian Wang, Peng Wang, Jian Sun, Adam Kortylewski, and Alan Yuille.
\newblock Voge: a differentiable volume renderer using gaussian ellipsoids for analysis-by-synthesis.
\newblock \emph{arXiv preprint arXiv:2205.15401}, 2022.

\bibitem[Wang et~al.(2023)Wang, Wang, and Agapito]{wang2023co}
Hengyi Wang, Jingwen Wang, and Lourdes Agapito.
\newblock Co-slam: Joint coordinate and sparse parametric encodings for neural real-time slam.
\newblock In \emph{Proceedings of the IEEE/CVF Conference on Computer Vision and Pattern Recognition}, pages 13293--13302, 2023.

\bibitem[Wu et~al.(2022)Wu, Lao, Jiang, Liu, and Zhao]{wu2022point}
Xiaoyang Wu, Yixing Lao, Li Jiang, Xihui Liu, and Hengshuang Zhao.
\newblock Point transformer v2: Grouped vector attention and partition-based pooling.
\newblock \emph{arXiv preprint arXiv:2210.05666}, 2022.

\bibitem[Xu et~al.(2021{\natexlab{a}})Xu, Zhang, Dou, Zhu, Sun, and Pu]{rpvnet}
Jianyun Xu, Ruixiang Zhang, Jian Dou, Yushi Zhu, Jie Sun, and Shiliang Pu.
\newblock Rpvnet: A deep and efficient range-point-voxel fusion network for lidar point cloud segmentation.
\newblock In \emph{IEEE/CVF International Conference on Computer Vision}, pages 16024--16033, 2021{\natexlab{a}}.

\bibitem[Xu et~al.(2021{\natexlab{b}})Xu, Zhang, Wei, Lin, Cao, Hu, and Bai]{xu2021simple}
Mengde Xu, Zheng Zhang, Fangyun Wei, Yutong Lin, Yue Cao, Han Hu, and Xiang Bai.
\newblock A simple baseline for zero-shot semantic segmentation with pre-trained vision-language model.
\newblock \emph{arXiv preprint arXiv:2112.14757}, 2021{\natexlab{b}}.

\bibitem[Xu et~al.(2023)Xu, Cong, Yao, Chen, Hou, Zhu, He, Yu, and Ma]{xu2023human}
Yiteng Xu, Peishan Cong, Yichen Yao, Runnan Chen, Yuenan Hou, Xinge Zhu, Xuming He, Jingyi Yu, and Yuexin Ma.
\newblock Human-centric scene understanding for 3d large-scale scenarios.
\newblock In \emph{Proceedings of the IEEE/CVF International Conference on Computer Vision}, pages 20349--20359, 2023.

\bibitem[Yan et~al.(2023)Yan, Qu, Wang, Xu, Wang, Zhao, and Li]{yan2023gs}
Chi Yan, Delin Qu, Dong Wang, Dan Xu, Zhigang Wang, Bin Zhao, and Xuelong Li.
\newblock Gs-slam: Dense visual slam with 3d gaussian splatting.
\newblock \emph{arXiv preprint arXiv:2311.11700}, 2023.

\bibitem[Yan et~al.(2022)Yan, Gao, Zheng, Zheng, Zhang, Cui, and Li]{XuYan20222DPASS2P}
Xu Yan, Jiantao Gao, Chaoda Zheng, Chaoda Zheng, Ruimao Zhang, Shenghui Cui, and Zhen Li.
\newblock 2dpass: 2d priors assisted semantic segmentation on lidar point clouds.
\newblock In \emph{ECCV}, 2022.

\bibitem[Yeshwanth et~al.(2023)Yeshwanth, Liu, Nie{\ss}ner, and Dai]{yeshwanth2023scannet++}
Chandan Yeshwanth, Yueh-Cheng Liu, Matthias Nie{\ss}ner, and Angela Dai.
\newblock Scannet++: A high-fidelity dataset of 3d indoor scenes.
\newblock In \emph{Proceedings of the IEEE/CVF International Conference on Computer Vision}, pages 12--22, 2023.

\bibitem[Yin et~al.(2024)Yin, Shen, Chen, Li, Yang, Frossard, and Wang]{yin2024fusion}
Junbo Yin, Jianbing Shen, Runnan Chen, Wei Li, Ruigang Yang, Pascal Frossard, and Wenguan Wang.
\newblock Is-fusion: Instance-scene collaborative fusion for multimodal 3d object detection.
\newblock In \emph{Proceedings of the IEEE/CVF Conference on Computer Vision and Pattern Recognition}, pages 14905--14915, 2024.

\bibitem[Yugay et~al.(2023)Yugay, Li, Gevers, and Oswald]{yugay2023gaussian}
Vladimir Yugay, Yue Li, Theo Gevers, and Martin~R Oswald.
\newblock Gaussian-slam: Photo-realistic dense slam with gaussian splatting.
\newblock \emph{arXiv preprint arXiv:2312.10070}, 2023.

\bibitem[Zhang and Ding(2021)]{zhang2021prototypical}
Hui Zhang and Henghui Ding.
\newblock Prototypical matching and open set rejection for zero-shot semantic segmentation.
\newblock In \emph{Proceedings of the IEEE/CVF International Conference on Computer Vision}, pages 6974--6983, 2021.

\bibitem[Zhu et~al.(2023)Zhu, Wang, Blum, Liu, Song, Pollefeys, and Wang]{zhu2023sni}
Siting Zhu, Guangming Wang, Hermann Blum, Jiuming Liu, Liang Song, Marc Pollefeys, and Hesheng Wang.
\newblock Sni-slam: Semantic neural implicit slam.
\newblock \emph{arXiv preprint arXiv:2311.11016}, 2023.

\bibitem[Zhu et~al.(2024)Zhu, Qin, Wang, Liu, and Wang]{zhu2024semgauss}
Siting Zhu, Renjie Qin, Guangming Wang, Jiuming Liu, and Hesheng Wang.
\newblock Semgauss-slam: Dense semantic gaussian splatting slam.
\newblock \emph{arXiv preprint arXiv:2403.07494}, 2024.

\bibitem[Zhu et~al.(2020)Zhu, Zhou, Wang, Hong, Ma, Li, Li, and Lin]{zhu2020cylindrical}
Xinge Zhu, Hui Zhou, Tai Wang, Fangzhou Hong, Yuexin Ma, Wei Li, Hongsheng Li, and Dahua Lin.
\newblock Cylindrical and asymmetrical 3d convolution networks for lidar segmentation.
\newblock \emph{arXiv preprint arXiv:2011.10033}, 2020.

\bibitem[Zhu et~al.(2021)Zhu, Zhou, Wang, Hong, Ma, Li, Li, and Lin]{zhu2021cylindrical}
Xinge Zhu, Hui Zhou, Tai Wang, Fangzhou Hong, Yuexin Ma, Wei Li, Hongsheng Li, and Dahua Lin.
\newblock Cylindrical and asymmetrical 3d convolution networks for lidar segmentation.
\newblock In \emph{IEEE/CVF Conference on Computer Vision and Pattern Recognition}, pages 9939--9948, 2021.

\bibitem[Zhu et~al.(2022)Zhu, Peng, Larsson, Xu, Bao, Cui, Oswald, and Pollefeys]{zhu2022nice}
Zihan Zhu, Songyou Peng, Viktor Larsson, Weiwei Xu, Hujun Bao, Zhaopeng Cui, Martin~R Oswald, and Marc Pollefeys.
\newblock Nice-slam: Neural implicit scalable encoding for slam.
\newblock In \emph{Proceedings of the IEEE/CVF Conference on Computer Vision and Pattern Recognition}, pages 12786--12796, 2022.

\bibitem[Zou et~al.(2024)Zou, Yang, Zhang, Li, Li, Wang, Wang, Gao, and Lee]{zou2024segment}
Xueyan Zou, Jianwei Yang, Hao Zhang, Feng Li, Linjie Li, Jianfeng Wang, Lijuan Wang, Jianfeng Gao, and Yong~Jae Lee.
\newblock Segment everything everywhere all at once.
\newblock \emph{Advances in Neural Information Processing Systems}, 36, 2024.

\end{thebibliography}
}


\end{document}